\documentclass[11pt,a4paper]{article}
\usepackage[T1]{fontenc}
\usepackage{lmodern}
\usepackage[margin=24mm]{geometry}
\usepackage{microtype}
\usepackage{amsmath,amssymb}
\usepackage{graphicx}
\usepackage{array,booktabs,calc,longtable}
\usepackage[font=small,labelfont=bf,labelsep=period,hypcap=false]{caption}
\usepackage[super,sort&compress]{natbib}
\usepackage{xurl}
\usepackage[hidelinks,unicode]{hyperref}

\DeclareUnicodeCharacter{00B1}{\ensuremath{\pm}}
\DeclareUnicodeCharacter{00B2}{\textsuperscript{2}}
\DeclareUnicodeCharacter{00D7}{\ensuremath{\times}}
\DeclareUnicodeCharacter{0394}{\ensuremath{\Delta}}
\DeclareUnicodeCharacter{2265}{\ensuremath{\geq}}

\newsavebox{\HRScaption}
\newcommand{\HRSfigure}[2]{%
  \clearpage
  \begingroup
  \centering
  \sbox{\HRScaption}{\begin{minipage}{\linewidth}%
    \captionof{figure}{#2}\label{fig:#1}%
  \end{minipage}}%
  \includegraphics[width=\linewidth,%
    height=\dimexpr\textheight-\ht\HRScaption-\dp\HRScaption-2\baselineskip\relax,%
    keepaspectratio]{figures/figure#1.png}\par
  \noindent\usebox{\HRScaption}\par
  \endgroup
}

\hypersetup{pdftitle={A world model for longitudinal health-trajectory prediction in older adults across multinational cohorts},pdfauthor={Jingyi Wu, Chenghua Guo, Qianlin Zuo, Pengfei Li, Luxia Zhang}}
\begin{document}
\begin{center}
{\LARGE\bfseries A world model for longitudinal health-trajectory prediction in older adults across multinational cohorts\par}
\vspace{0.8em}
{\large Jingyi Wu\textsuperscript{1,2}, Chenghua Guo\textsuperscript{2}, Qianlin Zuo\textsuperscript{2}, Pengfei Li\textsuperscript{1,2,*}, Luxia Zhang\textsuperscript{1,3}\par}
\vspace{0.6em}
{\small
\textsuperscript{1}National Institute of Health Data Science, Peking University, Beijing 100191, China\par
\textsuperscript{2}Advanced Institute of Information Technology, Peking University, Hangzhou 311215, China\par
\textsuperscript{3}Center for Digital Health and Artificial Intelligence, Peking University First Hospital, Beijing 100034, China\par
\vspace{0.5em}
* Corresponding to Pengfei Li, \href{mailto:pfli@bjmu.edu.cn}{pfli@bjmu.edu.cn}\par
Jingyi Wu: \href{mailto:joywu@pku.edu.cn}{joywu@pku.edu.cn}; Chenghua Guo: \href{mailto:1610122718@pku.edu.cn}{1610122718@pku.edu.cn}; Qianlin Zuo: \href{mailto:qianlinzuo@163.com}{qianlinzuo@163.com}; Pengfei Li: \href{mailto:pfli@bjmu.edu.cn}{pfli@bjmu.edu.cn}; Luxia Zhang: \href{mailto:zhanglx@bjmu.edu.cn}{zhanglx@bjmu.edu.cn}\par}
\end{center}

\section*{Abstract}
\textbf{Objective:} Aging is characterized by continuously evolving health states, yet most existing prediction models focus on static health risk at isolated time points rather than explicitly modelling how health trajectories unfold over time. This study aims to develop a health world model capable of representing evolving health states and simulating their future trajectories, providing a foundation for more personalized and dynamic healthy-aging management.

\textbf{Methods:} The health world model was developed based on the Joint-Embedding Predictive Architecture framework, comprising four functional components: a state encoder, an action- and context-conditioned predictor, a stop-gradient target encoder, and clinical outcome heads. The model was trained using data from the US Health and Retirement Study (HRS) and externally validated in two additional national aging cohorts: the English Longitudinal Study of Ageing (ELSA) and the China Health and Retirement Longitudinal Study (CHARLS). Clinical outcomes of mortality and functional decline, including activities of daily living (ADL) worsening and instrumental activities of daily living (IADL) worsening, were evaluated. Model performance was assessed across two complementary dimensions in an open-loop setting: discrimination and calibration in characterizing clinically meaningful risk, and validity in simulating plausible longitudinal trajectories under specified actions.

\textbf{Results:} In the held-out HRS test set, the model maintained strong discrimination across extended simulation horizons, with AUROC values for mortality, ADL worsening and IADL worsening decreasing from 0.89, 0.93 and 0.93 at the first horizon to 0.84, 0.84 and 0.85 at the fifth horizon, respectively. Predicted risks showed consistent monotonic stratification of subsequent clinical outcomes across horizons. The frozen model retained robust discrimination in both external cohorts, with mortality AUROC ranging from 0.881 to 0.867 in ELSA and from 0.801 to 0.796 in CHARLS across the evaluated horizons. Beyond passive forecasting, the model generated trajectories that preserved the expected pattern across predefined action regimes and responded directionally to counterfactual changes in lifestyle actions, consistent with observed associations.

\textbf{Conclusions:} This longitudinal health world model provides a framework for representing aging as an evolving health process, linking current health states to plausible future trajectories. Such models could offer a more longitudinal view of individual health and extend toward prospective, action-aware health intelligence, supporting more individualized health management.

\section*{Introduction}

Population aging is rapidly reshaping the global burden of health and social care. The global population aged 60 years and older is projected to increase from around 1.0 billion in 2020 to 1.4 billion by 2030 and 2.1 billion by 2050, when approximately 80\% of older people are expected to live in low- and middle-income countries\cite{ref1}. This demographic transition is accompanied by a growing burden of chronic disease, disability and multimorbidity. According to the Global Burden of Disease Study 2023, non-communicable diseases accounted for 1.80 billion disability-adjusted life-years globally in 2023, up from 1.45 billion in 2010\cite{ref2}. These changes are placing growing pressure on health systems to maintain health and functional independence over longer lifespans, underscoring the need to accurately characterize individuals\textquotesingle{} complex and evolving health states for precise, personalized care.

Medical artificial intelligence has made substantial progress in electronic health records (EHR)-based early disease diagnosis and predefined health-outcome risk estimation\cite{ref3,ref4}. Yet these models generally assume that human health is a static process, predicting what is likely to happen at a specified horizon; even when repeated observations are incorporated, longitudinal prediction remains outcome-centric, providing limited insight into how an individual\textquotesingle s health evolves over time\cite{ref5}. This assumption sits uneasily with the nature of aging itself, which is a dynamic and heterogeneous process in which multiple dimensions of health change and interact over time\cite{ref6}. Individuals with similar health profiles at one time point may consequently follow very different subsequent trajectories, as prior disease, behaviors, treatments, functional status and social circumstances evolve\cite{ref7,ref8}. Although recent longitudinal foundation models have begun to represent evolving patient states from large-scale clinical records, they remain focused on characterizing biological aging or frailty patterns, failing to model how individual health trajectories might evolve under counterfactual future conditions\cite{ref6,ref8}. Conversely, causal inference methods provide insight into how an outcome would differ under alternative interventions, yet a causal effect estimate is typically a summary of the contrast between interventions at a specified endpoint; it does not by itself represent the sequence of intermediate health states through which that effect may emerge, or constitute a complete longitudinal trajectory\cite{ref9}. A clinically meaningful representation of health therefore needs to capture how an individual's health state changes over time and how future trajectories may unfold under changing conditions\cite{ref9,ref10}. Such capabilities are important for earlier recognition of health deterioration, identify opportunities for timely intervention, and support more precise and personalized longitudinal care.

Digital twins and reinforcement learning frameworks have emerged to capture evolving patient states for dynamic decision support\cite{ref10,ref11,ref12}. Within this paradigm, world models are distinguished by their explicit ability to learn transition dynamics and simulate how a patient\textquotesingle s state evolves over successive future steps, without requiring new real-world observations at each step\cite{ref13,ref14}. Operationally, a health world model can represent an individual\textquotesingle s current state, propagate that state forward through time, and query alternative trajectories under hypothetical actions\cite{ref15,ref16}. This counterfactual rollout mechanism enables the evaluation of prospective intervention strategies prior to clinical deployment. Several studies have applied world models to dynamic treatment support, primarily for simulating acute disease evolution and treatment planning\cite{ref9,ref17,ref18,ref19}. For example, Yang et al.\cite{ref19} developed the Medical World Model (MeWM), which simulates disease dynamics from synthesized post-treatment tumor CT images to evaluate treatment efficacy and select an optimal action plan; in clinical evaluation, MeWM improved interventional physicians\textquotesingle{} decision-making. However, evidence remains limited on the development and validation of world models for capturing complex, heterogeneous aging trajectories across diverse populations, over long-term horizons, and under varying health actions.

Here, we developed a longitudinal health world model for older adults that learns evolving health-state representations and predicts future states through open-loop rollouts. Trained in the US Health and Retirement Study (HRS), the model was evaluated for multi-horizon prediction of mortality and functional decline and externally validated in aging cohorts from England (ELSA) and China (CHARLS). We further assessed its ability to simulate clinically plausible trajectories under counterfactual actions. These findings contribute to a broader shift in healthy-aging management from static prediction to adaptive, action-aware health intelligence supporting continuous, personalized decision-making.

\section*{Methods}

\subsection*{Data source and study population}

Model development and validation were conducted using three nationally representative longitudinal cohorts from the Health and Retirement Study family: HRS, the English Longitudinal Study of Ageing (ELSA), and the China Health and Retirement Longitudinal Study (CHARLS) (\hyperref[fig:1]{\textbf{Figure 1}}). The HRS was utilized for primary model development and internal validation, whereas ELSA and CHARLS served as external validation cohorts without model retraining. As sister surveys to the HRS, both ELSA and CHARLS share a common conceptual framework and biennial household-panel design, incorporating highly harmonized instruments across demographics, socioeconomic status, health behaviors, physician-diagnosed conditions, functional disability, cognition, and biomarkers\cite{ref20}.

The HRS is a nationally representative biennial panel of US adults aged ≥50 years and their spouses (available at \url{https://hrs.isr.umich.edu/}). Households are selected through a multistage area-probability sample, with oversampling of African American and Hispanic households and Florida residents. Sequential birth-year cohorts have been enrolled to maintain representation of the older population. Core interviews are administered approximately every two years, with proxy or exit interviews after death. Beginning in 2006 (wave 8), a rotating half-sample completed an enhanced face-to-face interview that added physical performance tests, anthropometrics, blood biomarkers and a psychosocial leave-behind questionnaire; we therefore started the modelling window at wave 8.

The ELSA is an ongoing, nationally representative prospective cohort study of adults aged ≥50 years living in England, designed to provide a longitudinal assessment of aging and later-life health (available at \url{https://www.elsa-project.ac.uk/}). The national baseline survey was conducted in 2002--2003, with participants followed at approximately 2-year intervals through 2020--2021 (waves 1--10). Data are collected through interviews and objective physical measurements, spanning demographics and family circumstances, physical and mental health and functioning, socioeconomic and employment status, and health-related behaviors and healthcare use. Repeated measurements across waves enable the assessment of changes in health status, behaviors and functional outcomes over time.

The CHARLS is a nationally representative prospective cohort designed to collect household- and individual-level information on middle-aged and older adults in China (available at \url{https://charls.pku.edu.cn/}). The study enrolls adults aged ≥45 years and was initiated with a national baseline survey in 2011--2012, with participants followed at approximately 2-year intervals through 2020--2021 (waves 1--5). CHARLS used a multistage, probability-proportional-to-size sampling design, covering 150 counties or districts and 450 village-level or urban-community units across China. The survey collects information spanning demographics and family structure, physical and cognitive health and functioning, socioeconomic and employment status, and health-related behaviors and healthcare use. The longitudinal design permits the examination of within-person changes in health and functional status over successive survey waves. Because the harmonized CHARLS panel contains five analyzable waves, a maximum of four consecutive wave-to-wave transitions was available for the present longitudinal analyses.

HRS protocols were approved by the University of Michigan Institutional Review Board (NO. HUM0061128); informed consent was obtained from all participants. ELSA was approved by the relevant National Health Service Research Ethics Committee, with informed consent obtained at each wave. Ethical approval for all CHARLS waves was granted by the Institutional Review Board at Peking University (household survey including anthropometrics, IRB00001052-11015; biomarker collection, IRB00001052-11014). The present analysis used de-identified public-use files under the surveys\textquotesingle{} data-use agreements and involved no additional participant contact.

\subsection*{Predictors and outcomes}

Across all three cohorts, the analysis unit was a person-wave transition: an interval from interview \emph{t} to the next observed interview or to a death-censored terminal interval. Participants were eligible if they contributed at least one such transition with observed baseline inputs. HRS participants were partitioned at the person level into training, validation and test sets (70\% / 15\% / 15\%; 78,262 / 16,530 / 16,740 transitions), so that no individual appeared in more than one split. ELSA and CHARLS were used only for external validation. Country-specific interviews are recoded onto a common variable schema based on the HRS instrument, via the Harmonized HRS, Harmonized ELSA and Harmonized CHARLS files released by the Gateway to Global Aging Data project, providing the basis for treating the three cohorts as aligned sources for model development and external validation. Residual differences in sampling frames, item availability and health-system context were retained as coverage flags and missingness masks rather than forced onto a single common scale. Further information on the data dictionary and harmonization rules is provided in the \textbf{Supplementary Data}.

For the primary HRS specification, we assembled a longitudinal sequence of dynamic states, static characteristics and actions at each person-wave. Dynamic state variables captured chronic conditions and self-rated health; functional status and mobility; sensory symptoms, pain and falls; cognition and psychosocial characteristics; physiological and performance measures; blood biomarkers; and healthcare use. Static context variables comprised demographics, early-life socioeconomic circumstances, household characteristics, income and wealth, and employment and insurance status. Actions represented potentially modifiable behaviors and treatments: cigarettes smoked per day; alcohol consumption (days per week and drinks per drinking day); binge drinking; frequency of vigorous, moderate and light physical activity (scored 0, never, to 4, most frequent); antihypertensive treatment; and diabetes medication. Medication actions were masked to unused (0) when the corresponding diagnosis was absent.

These variables were harmonized across cohorts through a series of preprocessing steps to construct the transition dataset used for training. ELSA and CHARLS items were first mapped to HRS variable names, units and value codes at the person-wave level. Continuous inputs were then standardized using training-set means and standard deviations only; validation, test and external-cohort data were transformed with this same HRS-derived scaler. Missing or inapplicable continuous and categorical fields were zero-filled and paired with a binary observation mask (0, missing or not applicable), and each categorical feature was represented as a learned embedding of dimension 16, with an additional index reserved for missingness. The inter-wave interval Δ\emph{t} (in years) was standardized and concatenated with the action embedding to accommodate irregular wave spacing. Transitions lacking eligible baseline inputs were excluded, as were outcome labels for which the subsequent wave was unobserved.

From these transitions, we defined three next-interval binary outcomes observed consistently across all three cohorts: all-cause death, worsening activities of daily living (ADL), and worsening instrumental ADL (IADL). Death was ascertained from survey mortality follow-up over the same interval. ADL (range 0--6) and IADL (range 0--5) worsening were each defined from the next-wave total score alone, coded 1 if that score reached ≥4 (severe impairment), 0 if the next wave was observed with a score \textless4, and missing otherwise.

\subsection*{Health world model development}

The health world model was developed based on the Joint-Embedding Predictive Architecture (JEPA) framework\cite{ref15}, to model how health states evolve over time and to simulate their future trajectories under specified actions (\hyperref[fig:1]{\textbf{Figure 1A}}). The model rests on two complementary principles: predicting future health states in a latent space, and rolling these latent states forward under explicit action and contextual conditions. This formulation allows the model to represent longitudinal health dynamics and to project potential future trajectories without requiring future covariates to be observed.

The model comprises four functional components: a state encoder, an action- and context-conditioned predictor, a stop-gradient target encoder, and clinical outcome heads. The state encoder integrates longitudinal dynamic health information to obtain a compact latent state (\emph{\textbf{z}\textsubscript{t}}); the predictor propagates this latent state forward conditional on static characteristics, actions and elapsed time; the target encoder, updated as an exponential moving average (EMA) of the state encoder, provides future latent representations for JEPA representation learning; and the clinical outcome heads translate the simulated latent trajectories into probabilities of death, ADL worsening and IADL worsening. Together, these components form an open-loop world model that learns latent health dynamics during training and can subsequently be queried to simulate alternative future trajectories under specified actions.

For each participant, the model received a longitudinal sequence of dynamic states (\emph{\textbf{s}}), static characteristics (\emph{\textbf{c}}), actions (\emph{\textbf{a}}), and elapsed time between observations (Δ\emph{t}). The dynamic history (\emph{\textbf{s}}) was first transformed into a sequence of state embeddings by the state encoder (\(Enc_{s}\)): each frame combined masked continuous variables, categorical embeddings and observation masks, and was projected into a 256-dimensional representation. A temporal transformer then integrated this sequence of state tokens to produce the current latent health state (\emph{\textbf{z}\textsubscript{t}}), representing the model\textquotesingle s learned summary of the participant\textquotesingle s evolving health state. This process can be parameterized as follows

\begin{equation}
\mathbf{z}_{t} = Temporal(Enc_{s}(\mathbf{s}_{1:t})).
\end{equation}

In parallel, static characteristics and actions were encoded separately. A static encoder produced a context embedding, while an action encoder represented the current action together with the elapsed time. These embeddings were provided to the predictor as conditions only; they were not incorporated into the target latent representation. This design allows the latent state to remain centered on dynamic health information while enabling the same state to be propagated under different action scenarios.

The central component of the world model is a conditioned predictor (\(f_{\theta}\)) that estimates the future latent health state from the current state and its contextual conditions:

\begin{equation}
{\widehat{\mathbf{z}}}_{t\mathbf{+}h} = f_{\theta}\left( {\widehat{\mathbf{z}}}_{t\mathbf{+}h - 1},\mathbf{c},\mathbf{a}_{t + h - 1},\mathrm{\Delta}t_{t + h - 1} \right),\ h = 1,\ \ldots,\ H.
\end{equation}

During training, the predictor was unrolled for three future horizons (\emph{H} = 3), with each predicted latent state feeding forward as the input to the next step. This open-loop rollout is central to the world-model formulation: once the initial state has been encoded, subsequent predictions require no newly observed covariates, allowing the model to simulate how a health trajectory would evolve under a specified sequence of actions.

JEPA training provided the learning signal for these latent transitions. A separate target branch, consisting only of the dynamic state encoder and temporal transformer, processed the subsequent observed sequence to obtain a target latent representation (\(\mathbf{z}_{t\mathbf{+}h}^{tgt}\)). The target branch was updated as an EMA of the dynamic state encoder and was detached from gradient propagation. The predictor was therefore trained to predict the representation of the future observed health state. The JEPA objective (\(L_{JEPA}\)) was defined as the mean-squared error (MSE) between the predicted and target latent representations across the rollout horizons. A variance regularizer was additionally applied at the first prediction step to discourage latent collapse. The training process works as follows

\begin{equation}
\mathbf{z}_{t\mathbf{+}h}^{tgt} = sg\lbrack{Temporal}_{EMA}\left( Enc_{s}^{EMA}\left( \mathbf{s}_{1:t + h} \right) \right)\rbrack,
\end{equation}

\begin{equation}
L_{JEPA}(h) = \mathbb{E}\lbrack {\|{\widehat{\mathbf{z}}}_{t\mathbf{+}h}\mathbf{-}\mathbf{z}_{t\mathbf{+}h}^{tgt}\|}_{2}^{2}\rbrack.
\end{equation}

The learned latent trajectory was linked to three clinically meaningful outcomes through independent clinical outcome prediction heads for death, ADL worsening, and IADL worsening. Clinical outcome losses (\(L_{clin}\)) were calculated at each rollout horizon using binary cross-entropy (BCE):

\begin{equation}
L_{clin}(h) = BCE(\widehat{p}\left( y_{t + h} \middle| {\widehat{\mathbf{z}}}_{t\mathbf{+}h} \right),y_{t + h}),
\end{equation}

where \(\widehat{p}\) denotes the predicted probability for a specific event (\emph{y}).

The overall optimization objective (\emph{L}) combined the horizon-specific JEPA loss, clinical outcome prediction loss, and latent-variance regularization:

\begin{equation}
L = \lambda_{JEPA}\sum_{h = 1}^{H}{w_{1}L_{JEPA}(h) +}\sum_{h = 1}^{H}{w_{2}L_{clin}(h)} + \lambda_{var}RELU(0.5 - std({\widehat{\mathbf{z}}}_{t\mathbf{+}1})),
\end{equation}

where \(\lambda\) and \emph{w} are weighting terms, and \(RELU(0.5 - std)\) denotes penalized standard deviation of one-step predicted latent.

Gradients were propagated through the online branch: the state encoder and temporal transformer, the static and action encoders, the predictor, and the clinical outcome heads. After each optimization step, only the target branch\textquotesingle s state encoder and temporal transformer were updated, as an EMA of their online counterparts; the target branch itself received no gradients. This asymmetric online--target design follows the JEPA principle, while preserving the action-conditioned transition mechanism that world-model simulation requires.

Once training was complete, the same architecture was used to simulate future trajectories from an observed clinical state. The participant\textquotesingle s dynamic history was first encoded, after which the predictor recursively generated future latent states under a specified sequence of actions. Alternative actions could then be introduced at a chosen horizon while holding static context, elapsed time and other actions constant. Each scenario was propagated forward independently, and the resulting latent trajectories were passed through the clinical heads to estimate the corresponding future outcomes.

Models were optimized using Adam, with a learning rate of 3×10\textsuperscript{-5}, weight decay of 5×10\textsuperscript{-4}, gradient clipping at 5, and a batch size of 256, trained for up to 40 epochs. Because death was a rare event within each interval, these intervals were oversampled during training to improve outcome learning. Model performance was evaluated every two epochs, and early stopping was based on a validation metric combining JEPA and clinical losses with equal weighting, using a patience of six evaluations and a minimum improvement threshold of 0.001. Detailed model weights for the developed health world model have been made publicly available through GitHub (\url{https://github.com/wujingyipku/Health-World-Model}).

\subsection*{Performance evaluation}

Model performance was evaluated across two complementary dimensions in an open-loop setting: discrimination and calibration abilities in characterizing clinically meaningful risk, and validity in simulating plausible longitudinal trajectories under specified actions (\hyperref[fig:1]{\textbf{Figure 1B}}). Internal evaluation used the person-level held-out HRS test set, whereas external evaluation applied the frozen HRS checkpoint directly to ELSA and CHARLS without updating any model parameters. Variables unavailable in a target cohort were treated as missing and assigned an observation mask of 0. At each horizon, the model generated next-interval predictions for death, ADL worsening and IADL worsening through the same open-loop mechanism used during training, recursively feeding each predicted latent state into the predictor rather than incorporating newly observed covariates.

We first assessed the discrimination of open-loop clinical predictions using the area under the receiver operating characteristic curve (AUROC), sensitivity at a probability threshold of 0.5, and sensitivity at a fixed specificity of ≥0.8. To distinguish longitudinal state forecasting from direct outcome prediction, we compared the world model against horizon-specific multilayer perceptron (MLP-h) models, each trained independently to predict the outcome at \emph{t}+\emph{h} directly from the same baseline feature vector available to the world model at the origin of the rollout. Relative performance was quantified as the ratio of the world-model metric to the corresponding MLP-h metric at each horizon. Each MLP-h comprised two hidden layers (128 and 64 units, ReLU activation), an L2 penalty of 10\textsuperscript{-4}, a batch size of 256 and a learning rate of 10\textsuperscript{-3}, with early stopping; class imbalance was addressed by replicating positive training observations up to the same maximum positive-weight equivalent of 10 used for the world model.

We next assessed the clinical relevance of the predicted risks by evaluating their calibration and their ability to stratify individuals by future adverse outcomes, with Kaplan-Meier (KM) curves used to compare death-free survival across predicted-risk strata. Participants were grouped into low-, intermediate- and high-risk tertiles according to their open-loop predicted probabilities at each horizon. Calibration was assessed by comparing mean predicted risk within each tertile against the corresponding observed event frequency, while stratification was assessed via KM death-free survival curves estimated according to risk tertiles defined at \emph{h} = 1, 2 and 3.

To assess whether the world model could reproduce clinically plausible trajectories beyond passive outcome prediction, we evaluated its behaviors under open-loop action rollouts, considering both static action regimes and counterfactual action changes. For static-regime analyses, participants were grouped according to action codes sustained across the first two observed waves, and the model generated trajectories following each participant\textquotesingle s observed action sequence. To ensure robust comparisons between empirical and simulated outcomes, we compared the observed and simulated age-standardized mortality risks using four baseline age bands (\textless60, 60--69, 70--79, and ≥80 years). For counterfactual analyses, we examined whether empirically observed action changes from a common baseline state produced coherent differences in the simulated trajectories. We restricted this analysis to participants who presented at a low or untreated level at the first horizon, classifying them according to any action changes occurring during the subsequent four horizons. To specifically simulate risk, we applied a pre-specified intervention at the second horizon for this baseline population, after which we rolled the model forward in an open loop to project the long-term effects of the new action regime. To capture diverse clinical domains, we selected three representative actions for both analyses: light physical activity (categorized as low {[}code 0{]}, medium {[}codes 1--2{]}, or high {[}codes 3--4{]}), smoking intensity (0, 1--20, or ≥20 cigarettes per day), and antihypertensive treatment status.

Finally, to test whether cross-cohort transfer depended on the availability of specific input features, we conducted an ablation analysis. For each target cohort, a restricted world model was retrained using the same HRS training, validation and test splits and the same hyperparameters, after removing variables unavailable in the target survey. Performance of the restricted models was then compared with that of the full model to assess the extent to which cross-cohort transfer was attributable to differences in available variables, rather than to the learned longitudinal dynamics.

\section*{Results}

\subsection*{Patient characteristics}

The analysis comprised 29,585 HRS participants (111,532 person-wave transitions; index waves 8--13), 17,003 ELSA participants (77,900 transitions; waves 1--9) and 23,927 CHARLS participants (72,952 transitions; waves 1--4) (\hyperref[tab:s1]{\textbf{Table S1}}). Mean follow-up durations were 7.7 ± 3.9, 10.2 ± 5.9, and 7.2 ± 2.3 years for HRS, ELSA, and CHARLS, respectively. At the index visit, CHARLS participants were generally younger (mean age 56.6 years vs. 63.0 in HRS and 61.6 in ELSA) and had a lower proportion of females (36.7\% vs. 57.6\% and 54.6\%). Socioeconomic and health profiles also varied across the cohorts. Tertiary education was most prevalent in HRS (21.6\%) and rare in CHARLS (1.8\%). Furthermore, the HRS cohort exhibited the highest mean body mass index (29.8 kg/m\textsuperscript{2}) and a substantially greater prevalence of cardiometabolic disease and multimorbidity compared to ELSA and CHARLS (hypertension: 51.4\% vs. 31.4\% and 15.5\%; diabetes: 19.1\% vs. 6.7\% and 4.0\%). Over the follow-up period, mortality occurred in 26.7\% of HRS, 16.2\% of ELSA, and 11.9\% of CHARLS participants. Additionally, at least one interval of ADL worsening was observed in 10.6\%, 8.4\%, and 7.9\% of participants, respectively, while IADL worsening occurred in 8.1\%, 5.4\%, and 8.2\%.

\subsection*{Open-loop prediction performance of clinical outcomes}

The world model learned a stable and highly utilized latent representation. At a single-step horizon (\emph{h} = 1), the model's predictor demonstrated significantly higher cosine similarity (0.958 vs. 0.574) and markedly lower latent MSE (0.092 vs. 0.855) compared to a latent persistence baseline. Crucially, this advantage was maintained during open-loop rollouts. While the predictor\textquotesingle s error remained consistently low (0.078--0.092), the persistence baseline error degraded rapidly, reaching 1.23 by \emph{h} = 5. This divergence increased the world model\textquotesingle s relative advantage from 0.11 to 0.067 over the simulation period (\hyperref[fig:s1]{\textbf{Figure S1}}).

Building upon this stable latent representation, the world model demonstrated strong predictive performance on the HRS test set for mortality, ADL worsening, and IADL worsening across extended simulation horizons (\hyperref[fig:2]{\textbf{Figure 2A-C)}}. Discriminative accuracy remained high from one to five horizons: AUROC for mortality was 0.89 at \emph{h} = 1 and 0.84 at \emph{h} = 5. Similar stability was observed for ADL worsening (AUROC 0.93 to 0.84) and IADL worsening (AUROC 0.93 to 0.85). Furthermore, sensitivity at a standard 0.5 probability threshold remained robust across this five-wave horizon. For intervals of \emph{h} = 1, 2, and 5, sensitivity values were 0.75, 0.78, and 0.90 for mortality; 0.77, 0.75, and 0.73 for ADL worsening; and 0.80, 0.75, and 0.66 for IADL worsening. This consistency was also evident when evaluating sensitivity at a strict specificity threshold (≥0.8), which recorded corresponding values of 0.82, 0.78, and 0.71 for mortality; 0.92, 0.86, and 0.74 for ADL worsening; and 0.90, 0.84, and 0.73 for IADL worsening.

The temporal robustness of the world model becomes particularly clear when compared to a horizon-specific MLP baseline. Although the MLP matched or slightly exceeded the world model\textquotesingle s discriminative performance at \emph{h} = 1 (AUROC of 0.98, 0.91, and 0.92 for mortality, ADL worsening, and IADL worsening, respectively), its accuracy rapidly deteriorated over longer horizons. By \emph{h} = 5, the MLP\textquotesingle s AUROC fell to 0.73, 0.79, and 0.76, whereas the world model maintained an AUROC of ≥0.84 across all three outcomes (representing a relative AUROC advantage of 1.07--1.14 at \emph{h} = 5). This degradation in the MLP was even more pronounced in event detection: sensitivity plummeted from 0.88 to 0.25 for mortality, 0.39 to 0.18 for ADL worsening, and 0.42 to 0.24 for IADL worsening. The world model\textquotesingle s sensitivity, by contrast, remained consistently high, yielding a relative advantage of 1.4-, 1.7-, and 1.5-fold at \emph{h} = 5 for sensitivity at specificity ≥0.8. Ultimately, these results indicate that the primary benefit of the world model over a one-step MLP lies in its sustained temporal robustness rather than an initial ranking advantage.

Beyond simple discrimination, the world model's open-loop predictions proved highly prognostic. Predicted risk tertiles were strictly monotonic and well-separated at every horizon. For instance, the mean predicted probabilities for ADL worsening at \emph{h} = 5 were 0.12, 0.26, and 0.68 across the respective tertiles, precisely tracking the realized clinical trajectories (\hyperref[fig:2]{\textbf{Figure 2D}}). Furthermore, KM survival curves for mortality, stratified by risk tertile assigned at a single baseline wave, demonstrated sharp and sustained separation (event-free survival of 0.95, 0.81, and 0.41 at \emph{h} = 1). This clear stratification persisted regardless of which of the first three horizons was used to derive the risk score (\hyperref[fig:2]{\textbf{Figure 2E}}), underscoring the model\textquotesingle s reliability for long-term clinical prognostication.

\subsection*{External validation across ELSA and CHARLS cohorts}

To determine whether the model\textquotesingle s open-loop clinical predictions transfer beyond the training population, we applied the HRS-trained world model to the ELSA (England; \hyperref[fig:3]{\textbf{Figure 3}}) and CHARLS (China; \hyperref[fig:4]{\textbf{Figure 4}}) cohorts. The predictive patterns replicated successfully in both external populations. Absolute discrimination remained robust, showing only a modest decline over extended horizons (for example, the AUROC for mortality ranged from 0.881 to 0.867 in ELSA, and from 0.801 to 0.796 in CHARLS). Crucially, the world model's relative advantage over a horizon-specific MLP baseline was at least as pronounced as in the HRS cohort. For mortality prediction, where the baseline model degraded to near-chance performance, the world model achieved a relative AUROC advantage of up to 1.45 in ELSA and 1.50 in CHARLS. Furthermore, sensitivity, risk-tertile separation, and long-term KM survival stratification were preserved in both cohorts. Together, these findings demonstrate that multi-wave clinical dynamics learned from a US-based cohort generalize effectively across diverse country-specific survey designs and follow-up structures without the need for local retraining.

To rigorously evaluate the source of this generalizability, we conducted ablation analyses by retraining the world model exclusively on the feature sets that overlapped with ELSA or CHARLS. When evaluated on the corresponding external cohorts without further refitting, these ablated world models retained their open-loop predictive advantage over the MLP baseline (\hyperref[fig:s2]{\textbf{Figure S2}} \textbf{and} \hyperref[fig:s3]{\textbf{S3}}). In the ELSA cohort, the ablated world model maintained high discrimination across all horizons (\emph{h} = 1--5), achieving AUROCs between 0.87 and 0.90, compared to 0.76 to 0.84 for the MLP. As the horizon lengthened, the model demonstrated its most substantial gains in overall sensitivity and sensitivity at a specificity of ≥0.8. In the CHARLS cohort (\emph{h} = 1--4), while the ablated model\textquotesingle s AUROC for ADL worsening narrowed closer to the baseline, it maintained a clear superiority for IADL worsening and mortality. Most notably, the world model\textquotesingle s sensitivity remained highly stable (0.68--0.76), whereas the MLP's sensitivity collapsed (0.14--0.23), resulting in up to a 5.3-fold relative sensitivity advantage by \emph{h} = 3. Consistent with the primary analysis, the open-loop risk tertiles in both ablation studies remained distinctly separated, and KM curves for mortality continued to reliably rank high- and low-risk patient trajectories.

\subsection*{Open-loop validation under static and counterfactual actions}

To assess whether the world model can simulate realistic clinical trajectories beyond passive forecasting, we evaluated its performance under fully open-loop static action regimes and counterfactual interventions. When operated in this configuration, the world model successfully preserved the expected dose-response ordering between static action regimes. This consistency was observed at every simulation horizon and across all three clinical outcomes. Taking light physical activity for example, the model predicted the highest risk for persistently inactive individuals and the lowest risk for those who were persistently active (for example, mortality risk at \emph{h} = 5 was approximately 0.75 versus 0.40; \hyperref[fig:5]{\textbf{Figure 5A-C}}).

Beyond static regimes, the model also responded appropriately to counterfactual behavioral changes. Forcing a simulated rollout to transition from a low- to a mid- or high-activity regime at \emph{h} = 2 attenuated the simulated risk relative to remaining inactive, aligning directionally with the observed empirical data (\hyperref[fig:5]{\textbf{Figure 5D-F}}). This simulated benefit emerged immediately following the intervention and persisted throughout the follow-up period. Notably, there was minimal additional benefit from transitioning to high- rather than mid-frequency activity, suggesting that the model\textquotesingle s counterfactual response is driven primarily by the shift out of inactivity rather than the absolute intensity of the new regime.

This responsive dynamic generalized across a broader suite of simulated interventions (\hyperref[fig:s4]{\textbf{Figure S4}} \textbf{and} \hyperref[fig:s5]{\textbf{S5}}). The world model successfully reproduced the expected harmful counterfactual response for smoking. Furthermore, it accurately distinguished on-treatment from off-treatment regimes for antihypertensive and antidiabetic medication use, responding dynamically to forced regimen switches. Across all evaluated interventions, the world model consistently preserved the observed rank order of the population strata, notwithstanding its overarching tendency to overestimate absolute risk.

We evaluated a representative clinical case from a held-out test participant (\hyperref[fig:1]{\textbf{Figure 1C}}), illustrating how the longitudinal world model supports individualized decision-making. This workflow operates across four core stages: representing an individual\textquotesingle s dynamic health state based on multi-source health data, forecasting their downstream health trajectory, simulating risk pathways under alternative actions, and informing personalized clinical recommendations. By enabling clinicians to systematically explore "what-if" scenarios before initiating treatment, the model provides an interactive computational environment for prospective risk assessment and decision support.

\section*{Discussion}

This study developed a longitudinal health world model for older adults that represents evolving health states and predicts their future trajectories through open-loop rollouts. The model showed consistent performance in predicting mortality and functional decline across multiple time horizons and retained its predictive capability when transferred to aging cohorts from England and China. Beyond outcome prediction, the model generated clinically plausible trajectories under specified action regimes and counterfactual action changes, demonstrating its potential to represent how health may evolve under different future conditions. These findings provide proof of concept that a longitudinal health world model can learn transferable representations of health-state dynamics and use them to generate future trajectories across aging populations, offering a potential foundation for anticipating individual health trajectories and supporting more precise and personalized longitudinal care.

Aging is inherently a dynamic process of health-state transitions. Older adults commonly experience concurrent changes in physical function, health behaviors, and healthcare use, with substantial heterogeneity in how these dimensions evolve over time. Conventional prediction models typically estimate predefined health outcomes at fixed horizons, and few have focused on characterizing and predicting the dynamic health trajectory for precise longitudinal care\cite{ref5}. Röösli et al.\cite{ref21} developed Delphi-2M, a generative transformer model trained on longitudinal health records from UK Biobank participants that learns disease trajectories and predicts the next disease event and its time of occurrence. Delphi-2M achieved AUROCs of up to 0.76 for next-disease prediction, decreasing to 0.70 at a 10-year horizon, and retained an average AUROC of 0.67 when externally evaluated in Danish individuals. Kraljevic et al.\cite{ref22} developed Foresight, a generative pretrained transformer that represents longitudinal EHR data as sequences of clinical concepts and uses autoregressive next-token prediction to generate future patient timelines, including diagnoses, medications, procedures and clinical findings; precision reached ranging 0.68-0.88 for predicting future clinical outcomes. Wang et al.\cite{ref8} developed LifeClock, an EHRFormer-based framework that learns longitudinal representations from temporally ordered clinical records and derives a biological-age measure reflecting individual deviations in health aging, achieving a biological-age prediction MAE of approximately 4.14 years in external validation. Despite these advances, existing approaches conceptualize longitudinal health primarily as a sequence of observable clinical events or as a summary measure of biological aging, and their core prediction tasks remain largely one-step-ahead. Our study extends this line of work by developing a longitudinal health world model that learns latent health-state transition dynamics and enables open-loop rollouts, allowing future health states to be recursively propagated over multiple time steps without requiring subsequent observations. This can help identify emerging deterioration before a discrete adverse event occurs, anticipate changes in functional status and mortality risk, and provide a longer temporal window for preventive intervention and care planning.

Several pilot studies explored to utilize world models to simulate acute disease evolution and inform dynamic treatment planning. For example, Mu et al.\cite{ref17} developed EHRWorld to model clinical care as a sequential state-transition process, updating patient states after each intervention and achieving a higher trajectory-prediction accuracy than general-purpose and medical large language models. Xu et al.\cite{ref18} developed medDreamer, which combines an adaptive feature-integration module for irregularly sampled EHR data with latent imagination and a two-stage policy-learning strategy. In offline evaluation, medDreamer achieved lower estimated mortality than clinician decisions for sepsis treatment (14.25\% versus 23.44\%). In oncology, MeWM developed by Yang et al.\cite{ref19} combines a treatment policy model with a generative tumor-dynamics model to synthesize post-treatment CT images and evaluate alternative treatment responses, increasing interventional physicians\textquotesingle{} F1-score for selecting the optimal treatment protocol by 13\%. CLARITY developed by Ding et al.\cite{ref23} models treatment-conditioned tumor trajectories in a structured latent space while incorporating temporal intervals and patient-specific clinical context, achieving a C-index of 0.786 for survival stratification. Our framework extends world modelling to longitudinal aging, where health states evolve gradually across multiple interacting domains and over substantially longer time horizons. Our model demonstrated greater open-loop prediction stability and accuracy than conventional machine-learning approaches, consistent with findings from previous world-model studies, and external validation across two large-scale national cohorts further supported its generalizability.

Importantly, this generalizability was observed despite substantial differences across the three national cohorts in demographic composition, social environments, and patterns of health and behaviors, providing a stringent setting for evaluating the transferability of learned health dynamics. The transportability of traditional prediction models optimized to capture endpoint-specific associations can be limited, when these associations are shaped by population-specific disease prevalence, demographic composition, and healthcare systems\cite{ref24}. In contrast, the temporal structure underlying health evolution, including the accumulation of disease burden, functional decline, and changes in behavioral and physiological states, may be more broadly shared across populations\cite{ref25}. By representing individuals in terms of evolving health states and predicting their subsequent transitions, world models may therefore capture features of health evolution that are less dependent on population-specific endpoint associations, providing a potentially more transferable basis for longitudinal prediction. This generalizability suggests that the learned health-state dynamics may retain relevance across geographically and contextually diverse aging populations.

Beyond improving prediction accuracy, representing health as an evolving state may offer clinical value that static risk prediction alone cannot provide, by situating an individual\textquotesingle s current risk within the broader trajectory of how their health is changing over time. This shift in perspective allows clinicians to distinguish not only patients who are currently at high risk, but also those whose health is deteriorating rapidly despite cross-sectional profiles similar to more stable individuals\cite{ref7}. Such a representation may also support longitudinal monitoring by continuously updating an individual\textquotesingle s health state as new observations become available, capturing the cumulative and interacting changes across multiple domains from which health outcomes often emerge\cite{ref26}. Because these models integrate changes in disease burden, functional status, and behaviors within a shared temporal representation, they may further enable more continuous management of multimorbidity and age-related decline. The clinical value of evolving health-state modeling therefore lies not in predicting a single future outcome, but in offering a dynamic view of where a patient stands in the course of their health trajectory, whether that trajectory is shifting, and when a change in clinical management may be warranted.

The ability of a world model to generate trajectories under alternative action scenarios may further shift health management from risk prediction toward prospective risk management\cite{ref14}. By conditioning health-state transitions on candidate actions, the model provides a computational framework for exploring how alternative behavioral or treatment scenarios may be associated with different subsequent trajectories. This enables comparison of potential preventive or therapeutic strategies based on their projected effects and trade-offs over time, thereby informing the prioritization, sequencing, and timing of interventions\cite{ref27,ref28}. For example, in the case study (\hyperref[fig:1]{\textbf{Figure 1}}), simulated trajectories suggested greater potential for risk reduction under increased physical activity and smoking reduction, supporting the prioritization of these interventions. At the population level, these capabilities could further inform the targeting of preventive programs by identifying which populations are more likely to benefit from specific preventive strategies. By comparing the projected health consequences of different intervention scenarios, such models could help identify which populations are likely to gain substantial improvements in future health trajectories from a given intervention and which are likely to see only limited benefit. This information could support more targeted allocation of preventive resources across both populations and intervention strategies, potentially reducing inefficient deployment of interventions and improving the cost-effectiveness of population health programs.

Several limitations should be considered. First, several limitations pertain to the underlying cohort data. Selection bias cannot be excluded, particularly because individuals who remain engaged across successive survey waves may differ systematically from those with intermittent participation or attrition. The two-year inter-wave interval inherent to the source studies may obscure shorter- and medium-term health fluctuations occurring between assessments. Self-reported measures also carry an inherent risk of reporting misclassification, a limitation common to survey-based aging cohorts. In addition, although HRS, ELSA, and CHARLS share a broadly consistent harmonization framework, fully equivalent semantic correspondence could not be achieved for every variable during cross-cohort translation, and residual definitional differences may have contributed to variation in model performance across cohorts. Second, the scope of the current model is constrained by the available state space, action space, and outcome definitions. The dynamic state representation, although extensive, does not capture potentially important determinants of health trajectories, such as laboratory measures, disease-specific biomarkers, and other clinical information that may capture additional dimensions of health status. Likewise, the action space considered in the present model represents only a subset of the potentially modifiable determinants of health trajectories. A broader, more granular action space, encompassing the full range of clinical, behavioral, and social interventions available in real-world care, will be needed to evaluate complex intervention strategies and their interactions across multimorbidity. Similarly, evaluation was restricted to mortality and functional deterioration, which represent important but incomplete dimensions of longitudinal health. Future work should examine whether the learned dynamics generalize to other clinically relevant trajectories; such evaluations would help establish whether the latent dynamics capture health evolution broadly, rather than trajectories specific to a limited set of outcomes.

Third, the world model itself has several inherent limitations. Recursive open-loop rollout may accumulate prediction error as the forecasting horizon lengthens: because predicted states are repeatedly fed back into the model, small transition errors at early steps can propagate to subsequent predictions and contribute to increasingly smoothed or uncertain trajectories\cite{ref9,ref29}. Moreover, because the model was trained and evaluated on observational cohort data, action-conditioned simulations do not establish the causal effects of behavioral or therapeutic change; unmeasured confounding, treatment selection, and time-varying confounding may all shape the transitions the model has learned\cite{ref30}. Specifically, the action-conditioned trajectories learned from observational data may be susceptible to reverse causation. For example, declining health may itself lead to inactivity or to the initiation of treatment, rather than the reverse. The causal interpretation and clinical utility of simulated interventions require prospective validation. Fourth, although external validation in ELSA and CHARLS provides evidence of cross-population generalizability, broader validation across additional countries, health systems, and demographic groups will be necessary to establish the portability of the learned health dynamics. Moreover, the latent health representation is not directly interpretable in clinical terms, and further work is needed to link latent transitions to specific physiological, functional, and behavioral processes. Finally, the clinical utility of the world model remains to be established prospectively. The present study evaluated predictive performance, external transportability, and model-based trajectory simulation using retrospective longitudinal cohorts, but did not test whether these outputs can improve clinical decision-making or patient outcomes in practice. Prospective studies will be needed to determine whether trajectory-based information can support the prioritization, timing, and selection of interventions, and ultimately improve the efficiency and effectiveness of preventive care.

In conclusion, this study demonstrates the potential of health world models to capture the dynamic and heterogeneous nature of aging across populations. By linking current health states to plausible future trajectories, such models could complement conventional risk prediction with a more longitudinal view of individual health and provide a basis for exploring alternative future scenarios. With further validation in prospective and clinically grounded settings, this approach may contribute to a shift towards more precise, personalized and adaptive management of health across the aging life course.

\section*{Statements \& Declarations}

\subsection*{Funding}

This work was supported by the National Natural Science Foundation of China (72622003, 72474010) and Hangzhou Natural Science Foundation (2025SZRJJ1809).

\subsection*{Competing Interests}

The authors have no relevant financial or non-financial interests to disclose.

\subsection*{Author Contributions}

\textbf{Jingyi Wu:} Conceptualization, Formal analysis, Validation, Writing original draft, Writing -- review \& editing. \textbf{Chenghua Guo:} Formal analysis, Writing -- review \& editing. \textbf{Qianlin Zuo:} Validation, Writing -- review \& editing. \textbf{Pengfei Li:} Conceptualization, Funding acquisition, Supervision, Writing -- review \& editing. \textbf{Luxia Zhang:} Supervision, Writing -- review \& editing.

\subsection*{Data availability}

The datasets generated and analyzed during the current study are publicly available upon application. Further enquiries can be directed to the corresponding author.

\subsection*{Code availability}

The analysis and figure-generation code reproducing the reported quantitative results is publicly available under the MIT license at \url{https://github.com/wujingyipku/Health-World-Model}. The repository includes the complete analysis and figure-generation pipeline, a top-level entry point for reproduction, a README file and a requirements file. No raw patient data are included. For transparency, the repository also provides the data-preparation script used to process the original protected records; however, executing this script requires authorized access to the underlying raw data.

\subsection*{Ethics declarations}

Ethical approval for all the Health and Retirement Study (HRS) waves was granted by the Institutional Review Board (IRB) at the University of Michigan (NO. HUM0061128). Written informed consent was obtained from participants prior to each interview.

Ethical approval for the English Longitudinal Study of Ageing (ELSA) was obtained from the South Central and South East Multicentre Research Ethics Committees, with subsequent approval provided by the London Multicentre Research Ethics Committee and other relevant local research ethics committees for later waves. The approval numbers were MREC/01/2/91 for Wave 1, MREC/04/2/006 for Wave 2, 05/MRE02/63 for Wave 3, 07/H0716/48 for Wave 4, 09/H0505/124 for Wave 5, 11/SC/0374 for Wave 6, 13/SC/0532 for Wave 7, 15/SC/0526 for Wave 8, 17/SC/0588 for Wave 9, and 21/SC/0030 for Wave 10. Written informed consent was obtained from all participants.

Ethical approval for all the CHARLS waves was granted by the IRB at Peking University. The IRB approval number for the main household survey, including anthropometrics, is IRB00001052-11015; the IRB approval number for biomarker collection, was IRB00001052-11014.

\HRSfigure{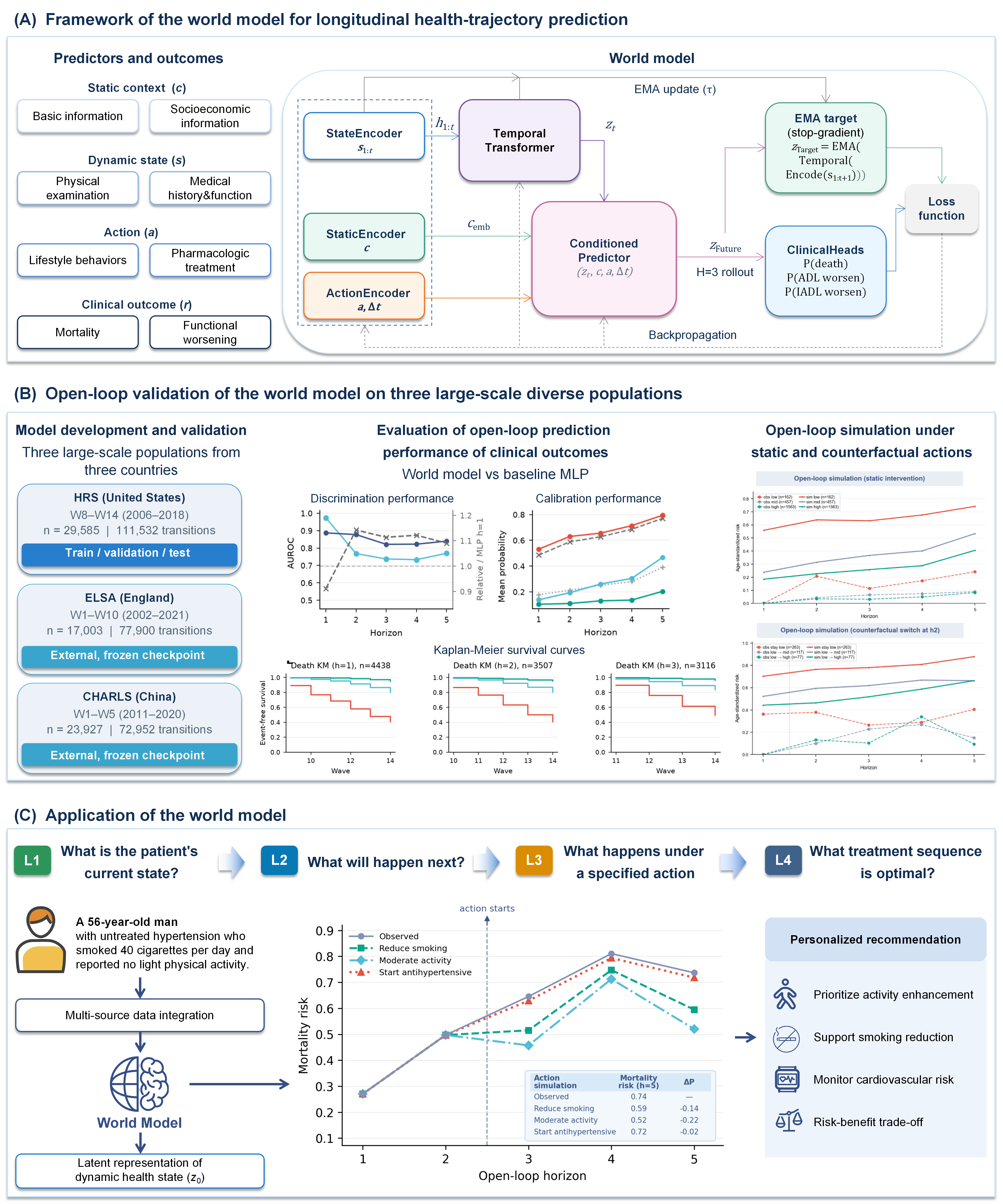}{\textbf{Study overview.} A, World model framework. The overall architecture designed for longitudinal health-trajectory prediction. B, Development and validation pipeline. Model training on the internal HRS cohort and external evaluation on two diverse populations (ELSA and CHARLS), assessing open-loop prediction and simulations under static and counterfactual actions. C, Clinical application case study. Demonstration of clinical decision support across four steps: health state representation, trajectory forecasting, counterfactual intervention simulation, and personalized health recommendations. CHARLS, China Health and Retirement Longitudinal Study; ELSA, English Longitudinal Study of Ageing; HRS, Health and Retirement Study.}

\HRSfigure{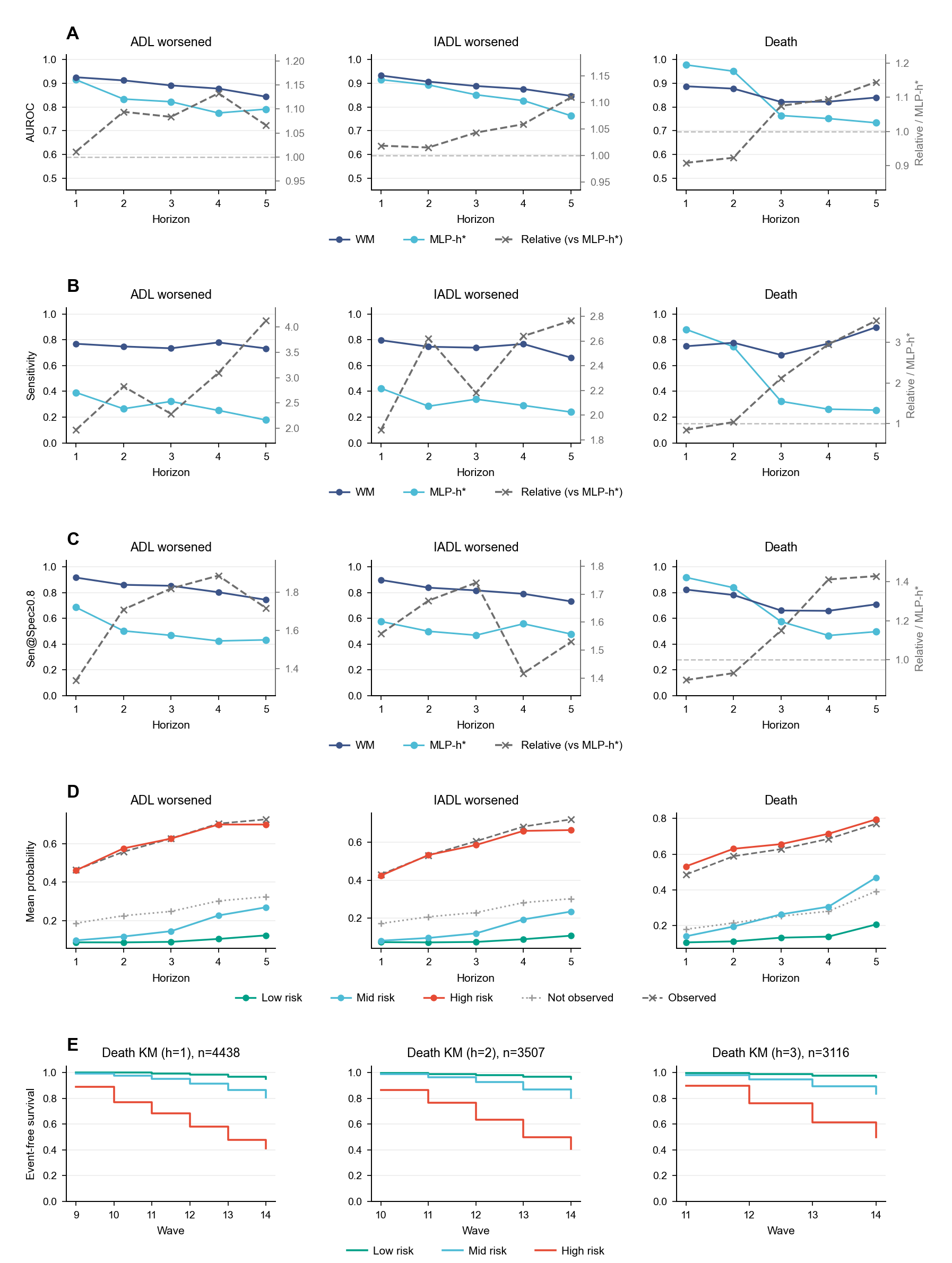}{\textbf{Open-loop prediction of clinical outcomes by the world model on the HRS test set.} Panels (A-C) show discrimination performance for ADL worsening, IADL worsening, and death across open-loop horizons h = 1-5 waves. Each row reports one metric: area under the receiver operating characteristic curve (AUROC; panel A), sensitivity (panel B), and sensitivity at specificity ≥0.8 (panel C). In each subplot, the left axis shows absolute performance of the world model (dark blue) and a MLP baseline (light blue), and the right axis shows relative performance (grey dashed line), defined as the ratio of world model at horizon \emph{h} to MLP; the horizontal dashed line marks parity (ratio = 1). Binary classifications use a probability threshold of 0.5. Panel (D) shows the mean open-loop predicted event probability by horizon, stratified by world model risk tertile (low, mid, high) and by whether the outcome was observed (dark grey dashed) or not observed (light grey dotted) on the realized trajectory. Panel (E) shows Kaplan-Meier estimates of event-free survival for death, stratified by open-loop risk tertile assigned at horizons \emph{h} = 1, 2 and 3 waves; n denotes the number of individuals in each horizon-specific cohort, and time is measured in HRS survey waves. ADL, activities of daily living; AUROC, area under the receiver operating characteristic curve; HRS, Health and Retirement Study; IADL, instrumental activities of daily living; MLP, multi-layer perceptron; WM, world model.}

\HRSfigure{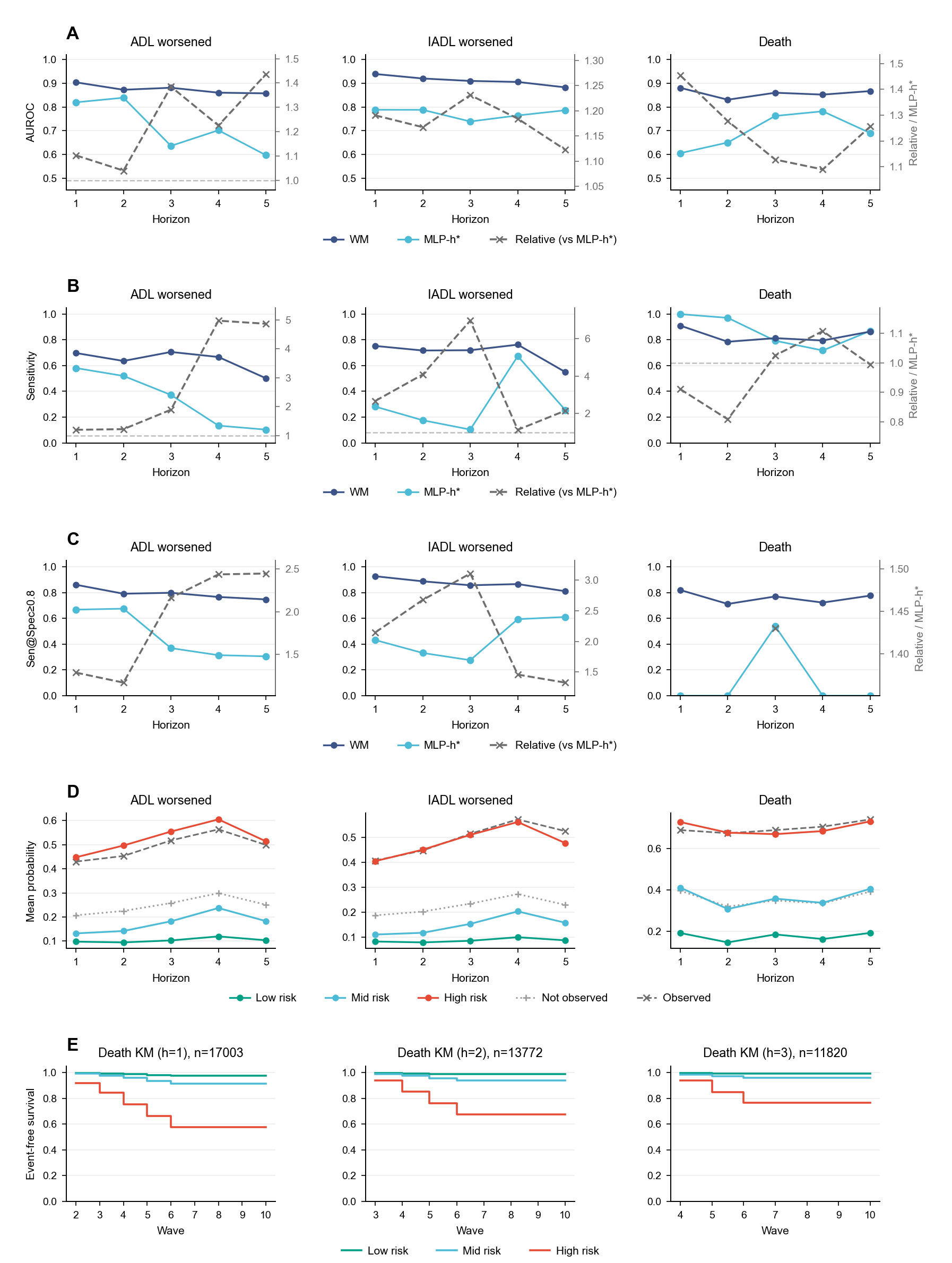}{\textbf{External validation of open-loop clinical outcome prediction in the ELSA cohort.} The world model was trained on HRS and evaluated without refitting on the English ELSA test set. Panels (A-C) show discrimination performance for ADL worsening, IADL worsening, and death across open-loop horizons \emph{h} = 1-5 waves. Each row reports one metric: area under the receiver operating characteristic curve (AUROC; panel A), sensitivity (panel B), and sensitivity at specificity ≥0.8 (panel C). In each subplot, the left axis shows absolute performance of the world model (dark blue) and a MLP baseline (light blue), and the right axis shows relative performance (grey dashed line), defined as the ratio of world model at horizon h to MLP; the horizontal dashed line marks parity (ratio = 1). Binary classifications use a probability threshold of 0.5. Panel (D) shows the mean open-loop predicted event probability by horizon, stratified by world model risk tertile (low, mid, high) and by whether the outcome was observed (dark grey dashed) or not observed (light grey dotted) on the realized trajectory. Panel (E) shows Kaplan-Meier estimates of event-free survival for death, stratified by open-loop risk tertile assigned at horizons \emph{h} = 1, 2 and 3 waves (n = 17,003, 13,772 and 11,820, respectively); time is measured in ELSA survey waves. ADL, activities of daily living; AUROC, area under the receiver operating characteristic curve; ELSA, English Longitudinal Study of Ageing; HRS, Health and Retirement Study; IADL, instrumental activities of daily living; MLP, multi-layer perceptron; WM, world model.}

\HRSfigure{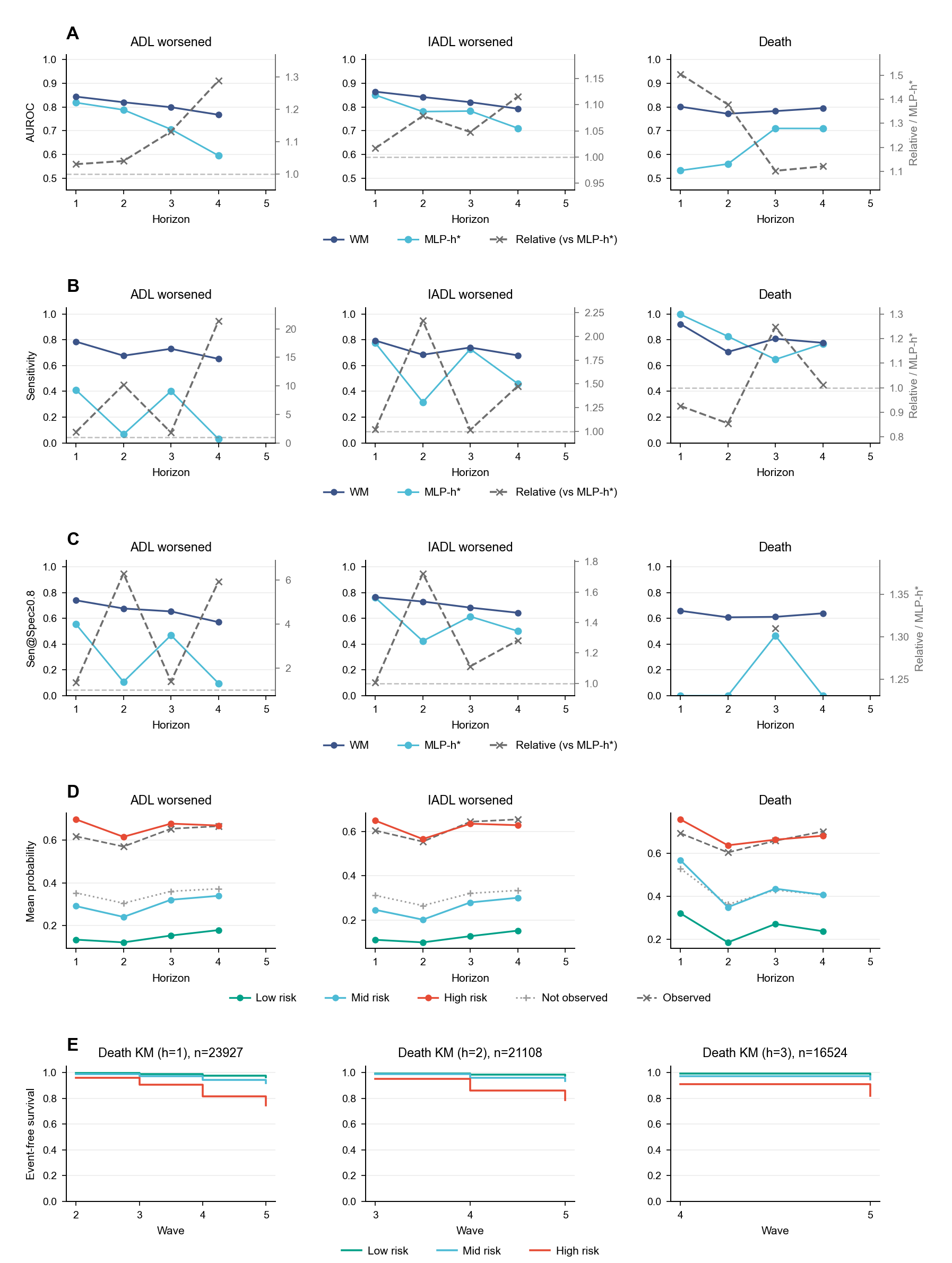}{\textbf{External validation of open-loop clinical outcome prediction in the CHARLS cohort.} The world model was trained on HRS and evaluated without refitting on the Chinese CHARLS test set. Panels (A-C) show discrimination performance for ADL worsening, IADL worsening, and death across open-loop horizons h = 1-4 waves. Each row reports one metric: area under the receiver operating characteristic curve (AUROC; panel A), sensitivity (panel B), and sensitivity at specificity ≥0.8 (panel C). In each subplot, the left axis shows absolute performance of the world model (dark blue) and a MLP baseline (ight blue), and the right axis shows relative performance (grey dashed line), defined as the ratio of world model at horizon \emph{h} to MLP; the horizontal dashed line marks parity (ratio = 1). Binary classifications use a probability threshold of 0.5. Panel (D) shows the mean open-loop predicted event probability by horizon, stratified by world model risk tertile (low, mid, high) and by whether the outcome was observed (dark grey dashed) or not observed (light grey dotted) on the realized trajectory. Panel (E) shows Kaplan-Meier estimates of event-free survival for death, stratified by open-loop risk tertile assigned at horizons \emph{h} = 1, 2 and 3 waves (n = 23,927, 21,108 and 16,524, respectively); time is measured in CHARLS survey waves. Horizons are truncated at \emph{h} = 4 in panels (A--D) because the harmonized CHARLS panel contributes only five analyzable follow-up waves, and hence at most four consecutive wave-to-wave transitions. ADL, activities of daily living; AUROC, area under the receiver operating characteristic curve; CHARLS, China Health and Retirement Longitudinal Study; HRS, Health and Retirement Study; IADL, instrumental activities of daily living; MLP, multi-layer perceptron; WM, world model.}

\HRSfigure{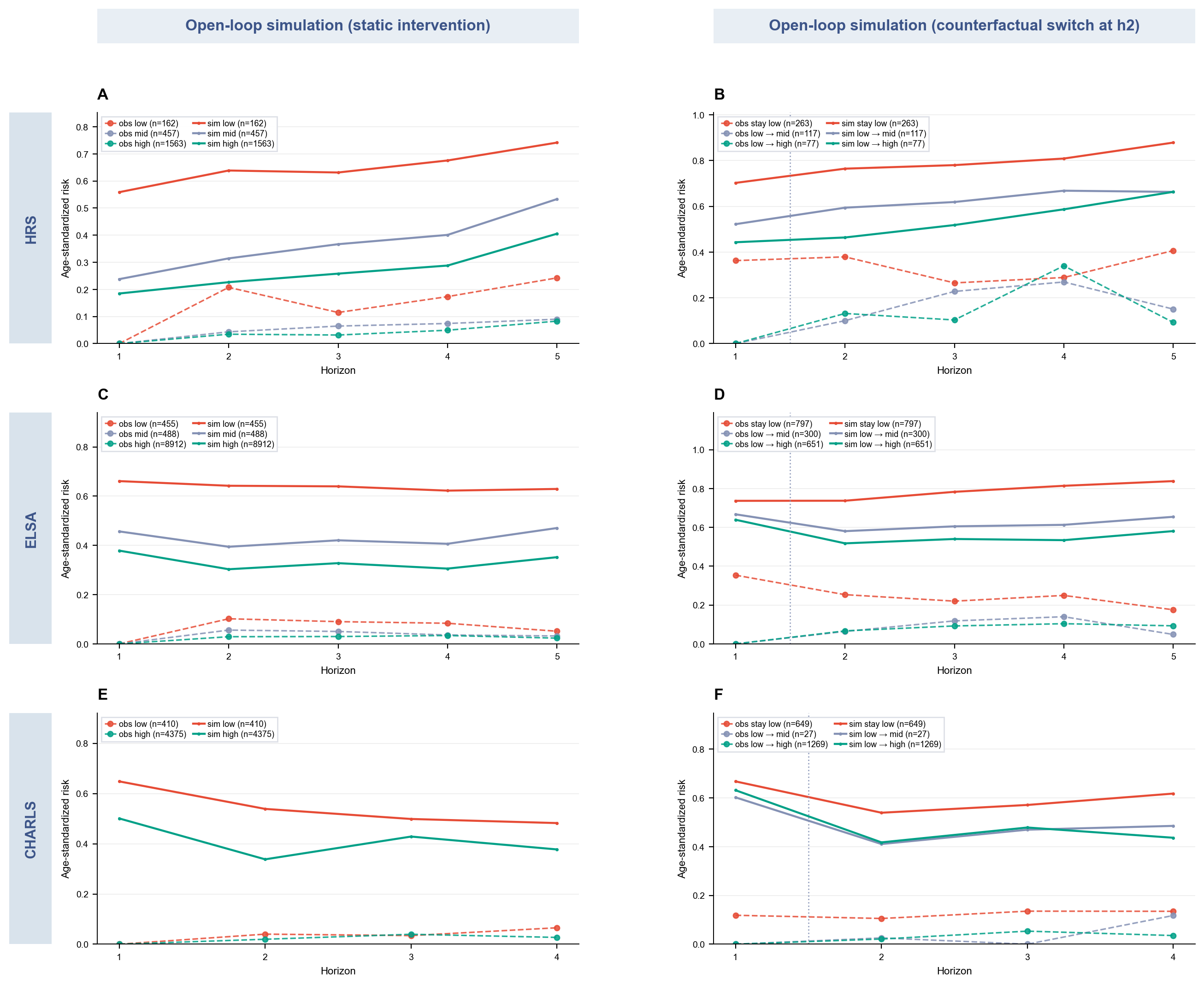}{\textbf{Open-loop validation of simulated age-standardized mortality risk under static and counterfactual light-activity-frequency exposure in HRS, ELSA and CHARLS.} The world model was trained on HRS and evaluated by open-loop simulation on held-out HRS individuals (A, B) and, without refitting, on the external ELSA (C, D) and CHARLS (E, F) cohorts. In the left column (A, C, E), individuals are stratified into three static, baseline-fixed light-activity-frequency regimes, defined from activity-frequency codes recorded at both of the first two waves; observed age-standardized risk (dashed lines, circles) is compared with world model-simulated risk (solid lines, dots) within each regime. In the right column (B, D, F), only individuals in the low regime at baseline are retained and further split into three groups by their empirically realized activity status at horizon 2: remained in the low regime, transitioned to the mid regime, and transitioned to the high regime; the world model rollout is counterfactually forced to switch to the corresponding regime at horizon 2, so that simulated and observed risk are compared under the same realized transition. CHARLS, China Health and Retirement Longitudinal Study; ELSA, English Longitudinal Study of Ageing; HRS, Health and Retirement Study.}

\clearpage
\section*{Supplementary Material}
\setcounter{figure}{0}
\renewcommand{\thefigure}{S\arabic{figure}}
\renewcommand{\theHfigure}{S\arabic{figure}}

\noindent\phantomsection\label{tab:s1}\textbf{Table S1. Baseline characteristics of study populations.}

\begin{longtable}[]{@{}
  >{\raggedright\arraybackslash}p{(\linewidth - 6\tabcolsep) * \real{0.4450}}
  >{\raggedleft\arraybackslash}p{(\linewidth - 6\tabcolsep) * \real{0.2000}}
  >{\raggedleft\arraybackslash}p{(\linewidth - 6\tabcolsep) * \real{0.1748}}
  >{\raggedleft\arraybackslash}p{(\linewidth - 6\tabcolsep) * \real{0.1802}}@{}}
\toprule\noalign{}
\begin{minipage}[b]{\linewidth}\raggedright
\textbf{Characteristic}
\end{minipage} & \begin{minipage}[b]{\linewidth}\centering
\textbf{HRS (United States)}
\end{minipage} & \begin{minipage}[b]{\linewidth}\centering
\textbf{ELSA (England)}
\end{minipage} & \begin{minipage}[b]{\linewidth}\centering
\textbf{CHARLS (China)}
\end{minipage} \\
\midrule\noalign{}
\endhead
\bottomrule\noalign{}
\endlastfoot
Participants, n & 29,585 & 17,003 & 23,927 \\
Person-wave transitions, n & 111,532 & 77,900 & 72,952 \\
Index interview wave range & 8--13 & 1--9 & 1--4 \\
Follow-up duration, years & 7.7 ± 3.9 & 10.2 ± 5.9 & 7.2 ± 2.3 \\
Follow-up duration, years, median (IQR) & 8.0 (4.0--12.0) & 10.0 (5.0--15.0) & 9.0 (5.0--9.0) \\
Age, years & 63.0 ± 11.8 & 61.6 ± 10.8 & 56.6 ± 10.5 \\
Female, n (\%) & 17,030 (57.6) & 9,282 (54.6) & 8,784 (36.7) \\
Tertiary education, n (\%) & 6,386 (21.6) & 2,590 (15.2) & 441 (1.8) \\
Married or partnered, n (\%) & 19,574 (66.2) & 12,454 (73.2) & 15,203 (63.5) \\
Self-rated health poor/fair, n (\%) & 9,205 (31.1) & 3,881 (22.8) & 8,727 (36.5) \\
ADL score (0--6) & 0.4 ± 1.1 & 0.4 ± 0.9 & 0.3 ± 0.9 \\
IADL score (0--5) & 0.3 ± 0.9 & 0.2 ± 0.6 & 0.3 ± 0.9 \\
CES-D score (0--8) & 1.6 ± 2.1 & 1.6 ± 2.0 & 2.1 ± 2.0 \\
Cognition score (0--27) & 15.1 ± 4.5 & 13.7 ± 5.0 & 12.3 ± 4.5 \\
Multimorbidity count & 1.8 ± 1.5 & 1.0 ± 1.1 & 0.9 ± 1.0 \\
BMI, kg/m² & 29.8 ± 6.3 & 28.3 ± 5.3 & 23.8 ± 3.9 \\
Hypertension, n (\%) & 15,218 (51.4) & 5,337 (31.4) & 3,714 (15.5) \\
Diabetes, n (\%) & 5,639 (19.1) & 1,142 (6.7) & 965 (4.0) \\
Heart disease, n (\%) & 5,670 (19.2) & 2,394 (14.1) & 1,818 (7.6) \\
Stroke, n (\%) & 2,146 (7.3) & 553 (3.3) & 402 (1.7) \\
Current smoker, n (\%) & 5,167 (17.5) & 2,992 (17.6) & 6,387 (26.7) \\
Death during follow-up, n (\%) & 7,897 (26.7) & 2,753 (16.2) & 2,842 (11.9) \\
ADL worsening event, n (\%) & 3,141 (10.6) & 1,429 (8.4) & 1,902 (7.9) \\
IADL worsening event, n (\%) & 2,410 (8.1) & 914 (5.4) & 1,966 (8.2) \\
\end{longtable}

ADL, activities of daily living; CES-D, Center for Epidemiologic Studies Depression Scale; CVD, cardiovascular disease; IADL, instrumental activities of daily living; IQR, interquartile range.

\HRSfigure{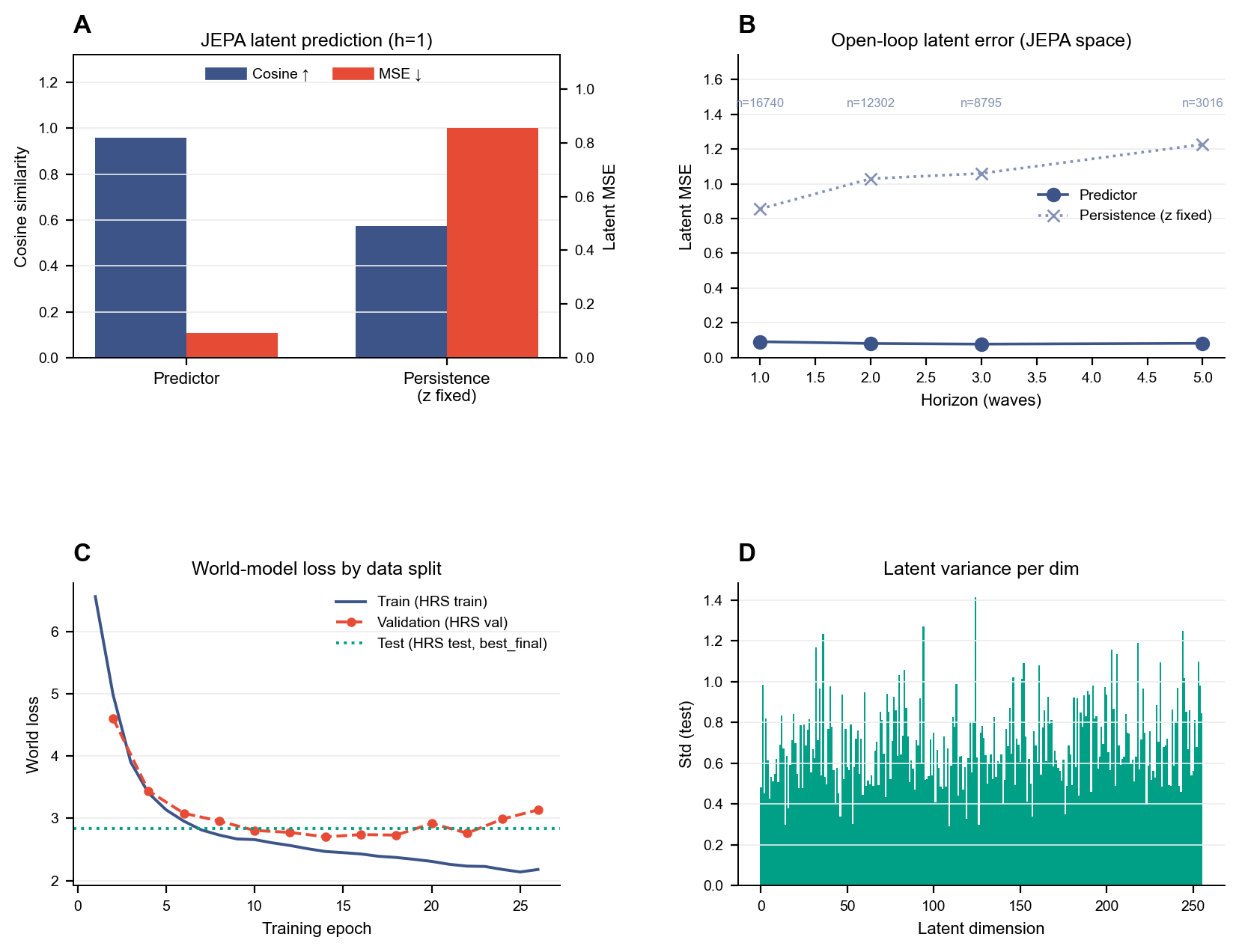}{\textbf{Evaluation of latent representation quality and training dynamics of the world model.} Panel (A) shows the one-step-ahead latent prediction on the HRS test set (\emph{h} = 1). Cosine similarity (left axis) and MSE (right axis) compare the JEPA predictor against a latent persistence baseline. Panel (B) shows the open-loop latent rollout error in JEPA embedding space across horizons h = 1-3 waves. Panel (C) shows training and validation loss curves over 25 epochs on HRS. Panel (D) shows standard deviation of each of the 256 latent dimensions on the HRS validation set, assessing dimensional collapse. HRS, Health and Retirement Study; JEPA, Joint-Embedding Predictive Architecture; MSE, mean squared error.}

\HRSfigure{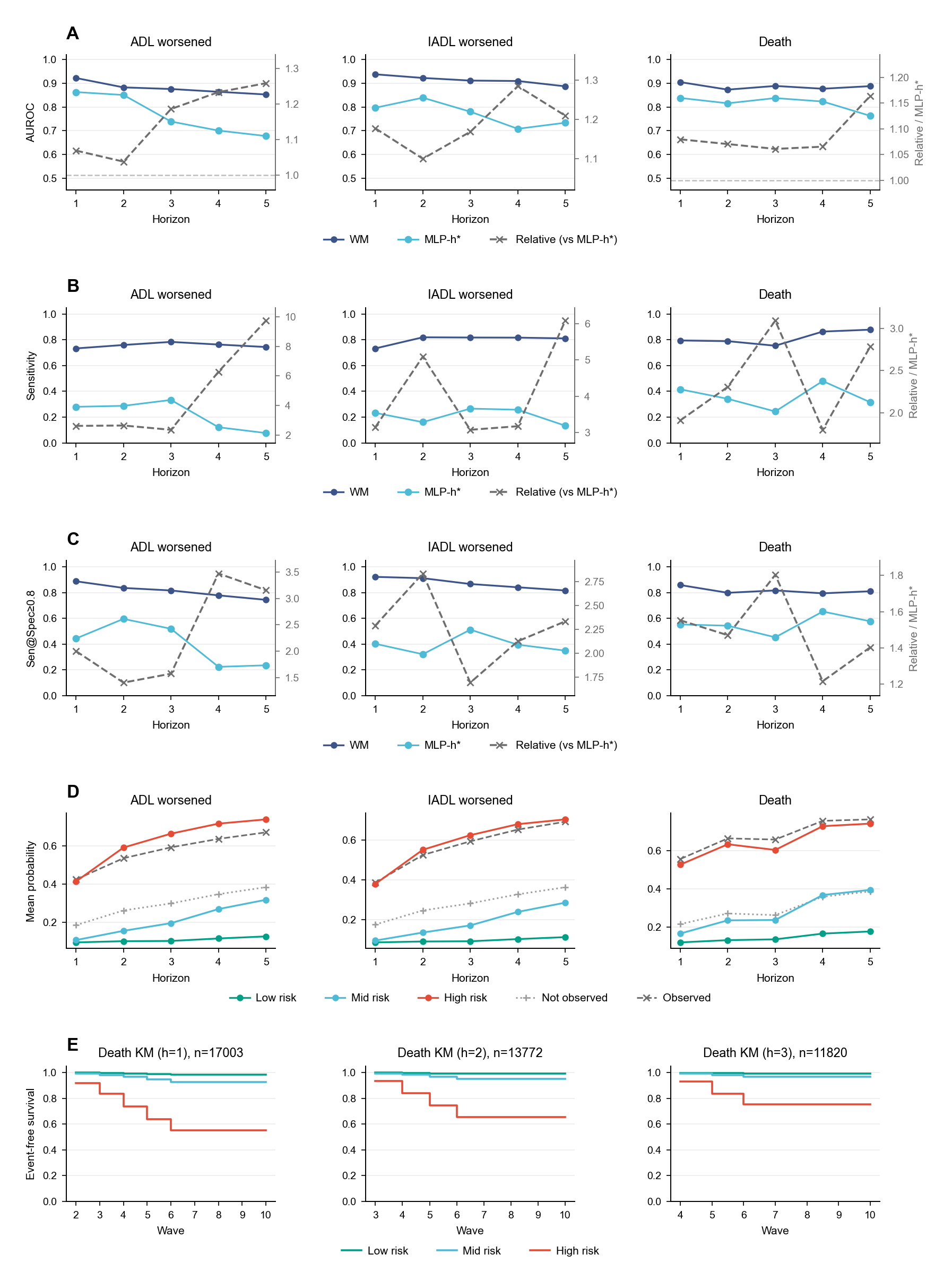}{\textbf{Ablation analysis of external validation of open-loop clinical outcome prediction in the ELSA cohort, using world model developed based on co-existing variables between HRS and ELSA cohorts.} The world model was trained on HRS and evaluated without refitting on the English ELSA test set. Panels (A-C) show discrimination performance for ADL worsening, IADL worsening, and death across open-loop horizons \emph{h} = 1-5 waves. Each row reports one metric: area under the receiver operating characteristic curve (AUROC; panel A), sensitivity (panel B), and sensitivity at specificity ≥0.8 (panel C). In each subplot, the left axis shows absolute performance of the world model (dark blue) and a MLP baseline (light blue), and the right axis shows relative performance (grey dashed line), defined as the ratio of world model at horizon \emph{h} to MLP; the horizontal dashed line marks parity (ratio = 1). Binary classifications use a probability threshold of 0.5. Panel (D) shows the mean open-loop predicted event probability by horizon, stratified by world model risk tertile (low, mid, high) and by whether the outcome was observed (dark grey dashed) or not observed (light grey dotted) on the realized trajectory. Panel (E) shows Kaplan-Meier estimates of event-free survival for death, stratified by open-loop risk tertile assigned at horizons \emph{h} = 1, 2 and 3 waves (n = 17,003, 13,772 and 11,820, respectively); time is measured in ELSA survey waves. ADL, activities of daily living; AUROC, area under the receiver operating characteristic curve; ELSA, English Longitudinal Study of Ageing; HRS, Health and Retirement Study; IADL, instrumental activities of daily living; MLP, multi-layer perceptron; WM, world model.}

\HRSfigure{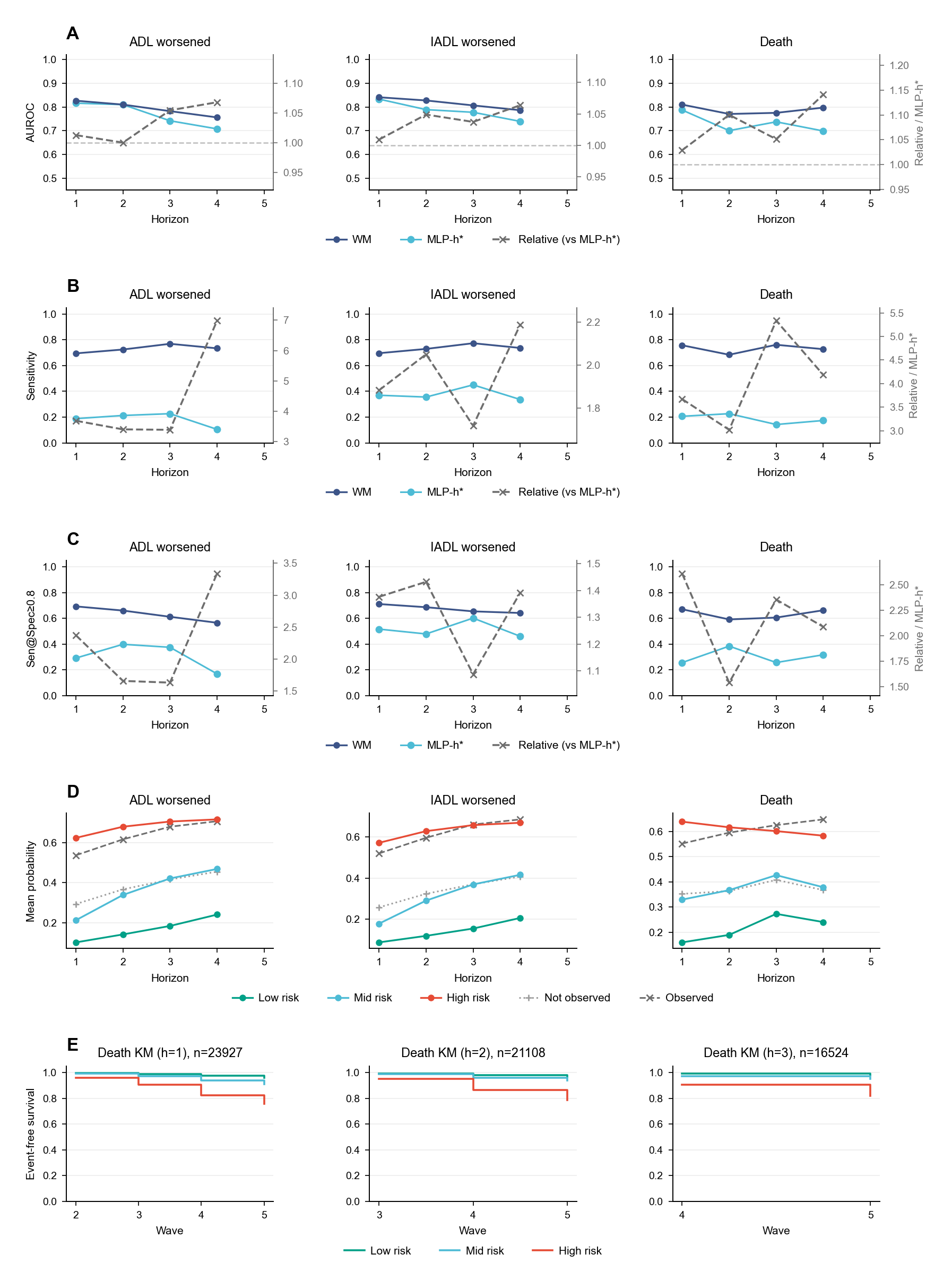}{\textbf{Ablation analysis of external validation of open-loop clinical outcome prediction in the CHARLS cohort, using world model developed based on co-existing variables between HRS and CHARLS cohorts.} The world model was trained on HRS and evaluated without refitting on the Chinese CHARLS test set. Panels (A-C) show discrimination performance for ADL worsening, IADL worsening, and death across open-loop horizons \emph{h} = 1-4 waves. Each row reports one metric: area under the receiver operating characteristic curve (AUROC; panel A), sensitivity (panel B), and sensitivity at specificity ≥0.8 (panel C). In each subplot, the left axis shows absolute performance of the world model (dark blue) and a MLP baseline (light blue), and the right axis shows relative performance (grey dashed line), defined as the ratio of world model at horizon \emph{h} to MLP; the horizontal dashed line marks parity (ratio = 1). Binary classifications use a probability threshold of 0.5. Panel (D) shows the mean open-loop predicted event probability by horizon, stratified by world model risk tertile (low, mid, high) and by whether the outcome was observed (dark grey dashed) or not observed (light grey dotted) on the realized trajectory. Panel (E) shows Kaplan-Meier estimates of event-free survival for death, stratified by open-loop risk tertile assigned at horizons \emph{h} = 1, 2 and 3 waves (n = 23,927, 21,108 and 16,524, respectively); time is measured in CHARLS survey waves. Horizons are truncated at \emph{h} = 4 in panels (A--D) because the harmonized CHARLS panel contributes only five analyzable follow-up waves, and hence at most four consecutive wave-to-wave transitions. ADL, activities of daily living; AUROC, area under the receiver operating characteristic curve; CHARLS, China Health and Retirement Longitudinal Study; HRS, Health and Retirement Study; IADL, instrumental activities of daily living; MLP, multi-layer perceptron; WM, world model.}

\HRSfigure{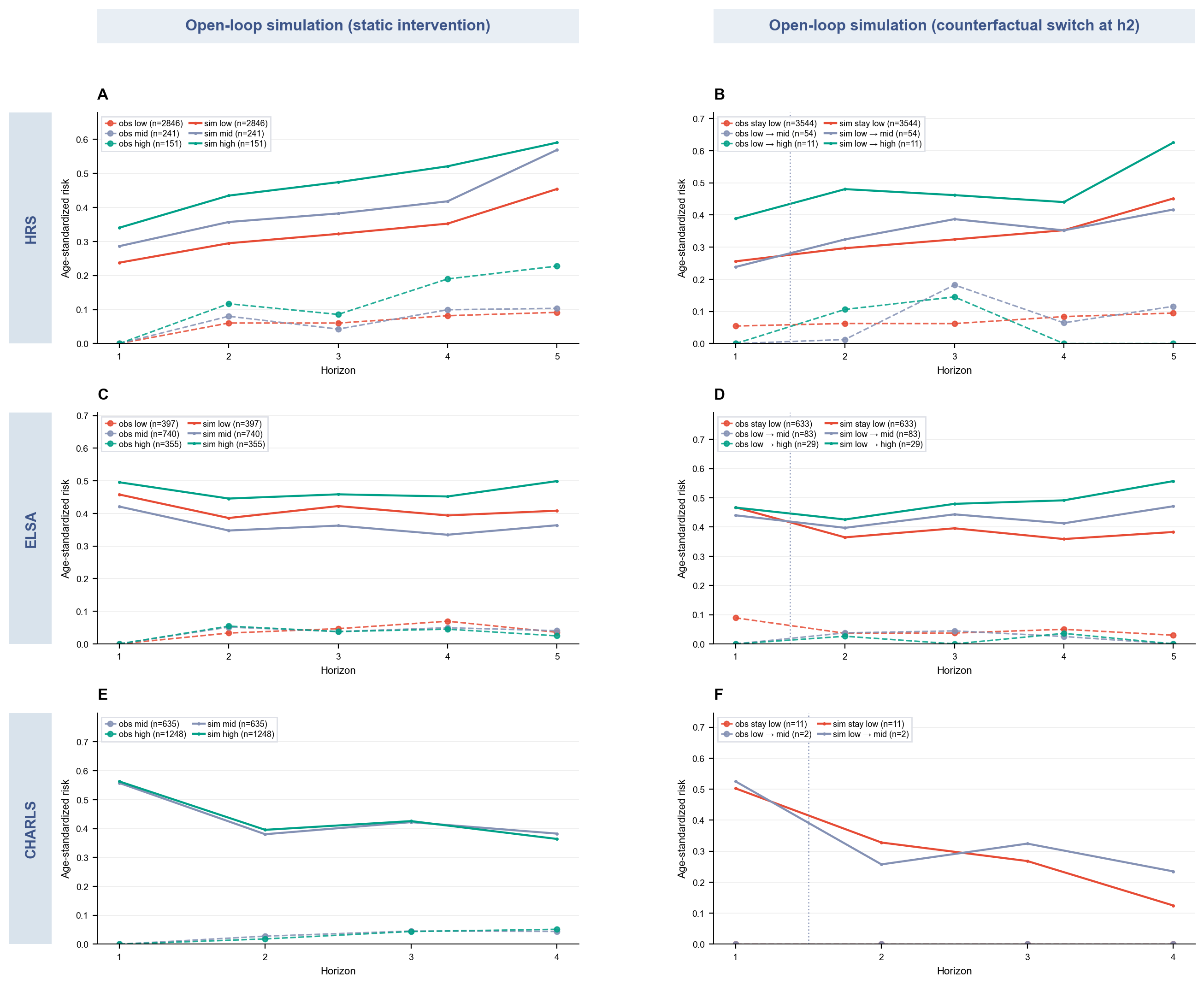}{\textbf{Open-loop validation of simulated age-standardized mortality risk under static and counterfactual smoking-intensity exposure in HRS, ELSA and CHARLS.} The world model was trained on HRS and evaluated by open-loop simulation on held-out HRS individuals (A, B) and, without refitting, on the external ELSA (C, D) and CHARLS (E, F) cohorts. In the left column (A, C, E), individuals are stratified into three static, baseline-fixed smoking-intensity regimes, defined from cigarettes smoked per day codes recorded at both of the first two waves; observed age-standardized risk (dashed lines, circles) is compared with world model-simulated risk (solid lines, dots) within each regime. In the right column (B, D, F), only individuals in the low regime at baseline are retained and further split into three groups by their empirically realized activity status at horizon 2: remained in the low regime, transitioned to the mid regime, and transitioned to the high regime; the world model rollout is counterfactually forced to switch to the corresponding regime at horizon 2, so that simulated and observed risk are compared under the same realized transition. CHARLS, China Health and Retirement Longitudinal Study; ELSA, English Longitudinal Study of Ageing; HRS, Health and Retirement Study.}

\HRSfigure{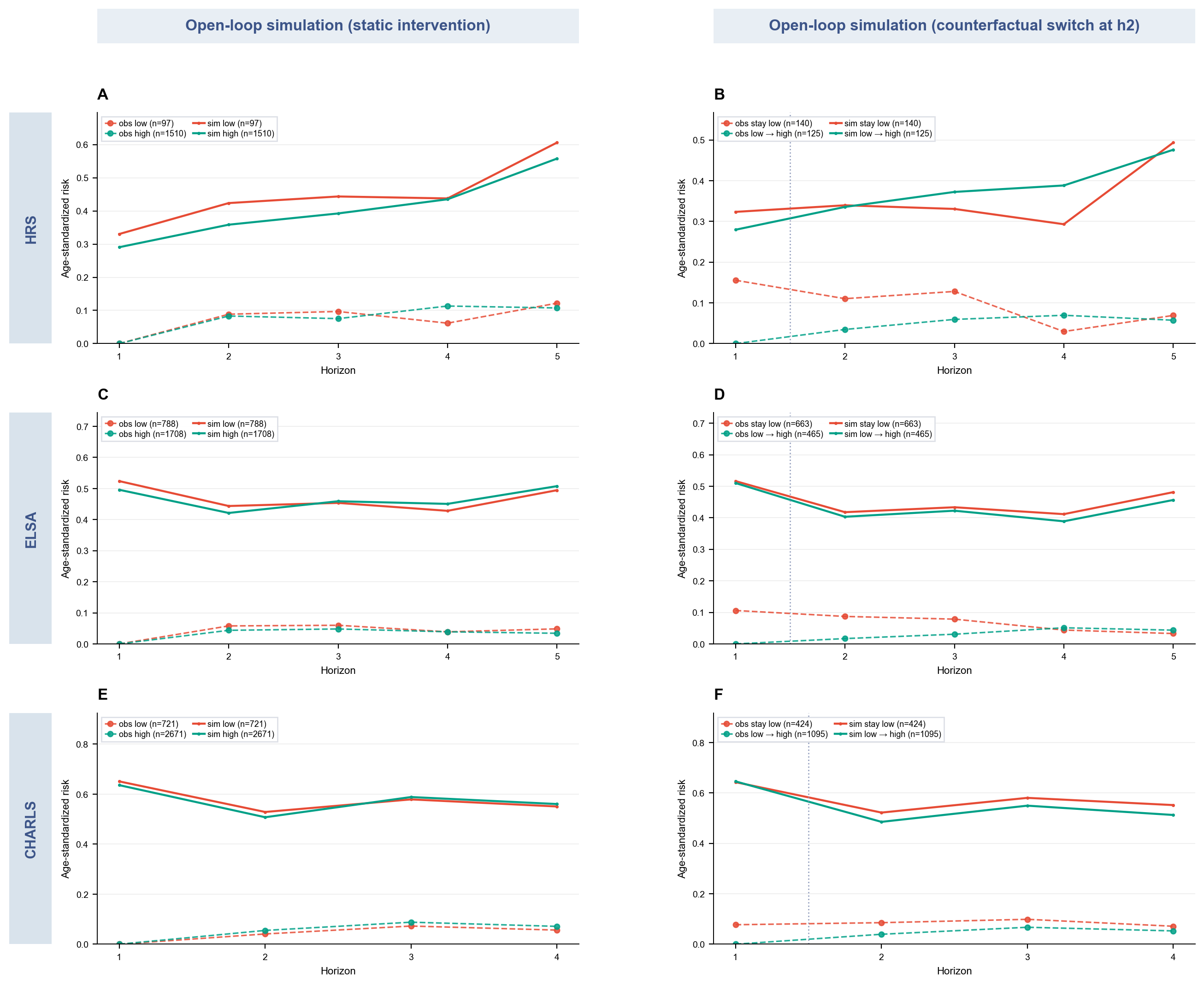}{\textbf{Open-loop validation of simulated age-standardized mortality risk under static and counterfactual antihypertensive medication exposure in HRS, ELSA and CHARLS.} The world model was trained on HRS and evaluated by open-loop simulation on held-out HRS individuals (A, B) and, without refitting, on the external ELSA (C, D) and CHARLS (E, F) cohorts. In the left column (A, C, E), individuals are stratified into two static, baseline-fixed antihypertensive medication regimes, defined from antihypertensive medication codes recorded at both of the first two waves; observed age-standardized risk (dashed lines, circles) is compared with world model-simulated risk (solid lines, dots) within each regime. In the right column (B, D, F), only individuals in the low regime at baseline are retained and further split into two groups by their empirically realized activity status at horizon 2: remained in the low regime and transitioned to the high regime; the world model rollout is counterfactually forced to switch to the corresponding regime at horizon 2, so that simulated and observed risk are compared under the same realized transition. CHARLS, China Health and Retirement Longitudinal Study; ELSA, English Longitudinal Study of Ageing; HRS, Health and Retirement Study.}

\end{document}